\documentclass[letterpaper]{article}

\usepackage{natbib,alifeconf}  
\usepackage{amsmath}
\usepackage{amsfonts}
\usepackage[backref=page]{hyperref}
\usepackage{url}
\usepackage[capitalize]{cleveref}
\usepackage{booktabs}
\usepackage{subcaption}
\usepackage{graphicx}
\usepackage{microtype}
\usepackage{comment}
\usepackage{float}
\usepackage{xcolor}         

\hypersetup{
  colorlinks   = true,              
  urlcolor     = blue,              
  linkcolor    = blue,              
  citecolor    = teal                
}

\usepackage{xspace}

\newcommand{\name}{JaxLife\xspace}

\title{\name: An Open-Ended Agentic Simulator}

\author{
    Chris Lu$^*$, Michael Beukman$^*$, Michael Matthews \and Jakob Foerster \\
    \mbox{} \\
    FLAIR, University of Oxford, United Kingdom \\
    $^*$Equal Contribution \\
    christopher.lu@exeter.ox.ac.uk, mbeukman@robots.ox.ac.uk
} 

\begin{document}

\maketitle

\begin{abstract}
    Human intelligence emerged through the process of natural selection and evolution on Earth. We investigate what it would take to re-create this process \textit{in silico}. While past work has often focused on low-level processes (such as simulating physics or chemistry), we instead take a more targeted approach, aiming to evolve \textit{agents} that can accumulate open-ended culture and technologies across generations.
    Towards this, we present \name: an artificial life simulator in which embodied agents, parameterized by deep neural networks, must learn to survive in an expressive world containing programmable systems. First, we describe the environment and show that it can facilitate meaningful Turing-complete computation. We then analyze the evolved emergent agents' behavior, such as rudimentary communication protocols, agriculture, and tool use. Finally, we investigate how complexity scales with the amount of compute used. We believe \name takes a step towards studying evolved behavior in more open-ended simulations.\footnote{Our code is available at \url{https://github.com/luchris429/JaxLife}.}
\end{abstract}

\section{Introduction}
Human capabilities, culture and intelligence have emerged from open-ended evolution on Earth \citep{darwin2023origin}. It follows that a multi-billion-year full-fidelity physics simulation of Earth could produce similarly-capable beings. However, such an endeavor is clearly computationally infeasible.
To reduce computational costs, one can reduce the fidelity of the simulation, raising the question of which components are necessary for the desired behavior.
To answer this, we must specify what behavior or capabilities we would like to potentially emerge from the simulation.

One such objective is to evolve agents that are capable of advanced reasoning and tool-use~\citep{parisi1997artificial}. After all, many of humanity's recent achievements involve mathematical reasoning and technical prowess. It may be the case that low-level control and perception---aspects that many simulations aim to reproduce~\citep{dittrich2001artificial,hutton2002evolvable}---are not necessary to evolve these capabilities. Indeed, even evolving morphologies, as many simulations do~\citep{sims1994Evolving,silveira1998modeling,spector2007division,bessonov2015morphology,pathak2019learning,alien}, may not be necessary for the evolution of advanced reasoning.

For this reason, we focus on the evolutionary advancements that make humans \textit{different} from other animals. 
Recent trends in anthropology focus on the idea of ``cultural accumulation'' \citep{henrich2015secret} as being the primary evolutionary origins of human intelligence. Cumulative culture is characterized by large amounts of social learning and the persistence and continuous advancement of shared knowledge. \citet{muthukrishna2018cultural} builds a simple computational model of the emergence of cultural accumulation and finds that it can be facilitated by a small bias towards social learning: If indeed this is humanity's key defining feature, it may not be difficult to replicate \textit{in silico}.

Our work is not the first to investigate the emergence of intelligent behavior. Prior works have modeled the emergence of cumulative culture through agent-based models~\citep[ABMs]{muthukrishna2018cultural, lu2022model}, which are high-level statistical models of agents. However, ABMs have not produced agents that are capable of advanced reasoning and instead model simplified high-level evolutionary dynamics. Similarly, other work in deep reinforcement learning has produced emergent social behaviors~\citep{johanson2022emergent}, communication~\citep{chaabouni2021emergent},  and tool-use~\citep{baker2019emergent} in games. Each of these settings suffers from the same failure mode: Their environments are not \textit{expressive enough} to produce truly open-ended expression. For example, agents in these environments cannot reasonably communicate about mathematics or build advanced machines. 

Our work aims to address this gap by allowing high-level agents to interact with and program composable robots that can express useful and meaningful Turing-complete behaviors. 
We present \name, an artificial life simulator capable of expressing meaningful, Turing-complete behaviors. 
Our simulator is written entirely in JAX~\citep{jax2018github}, meaning it can run on hardware accelerators and easily scale to multiple devices. 
Our contributions are as follows:

\begin{figure*}
\centering
\includegraphics[width=\linewidth]{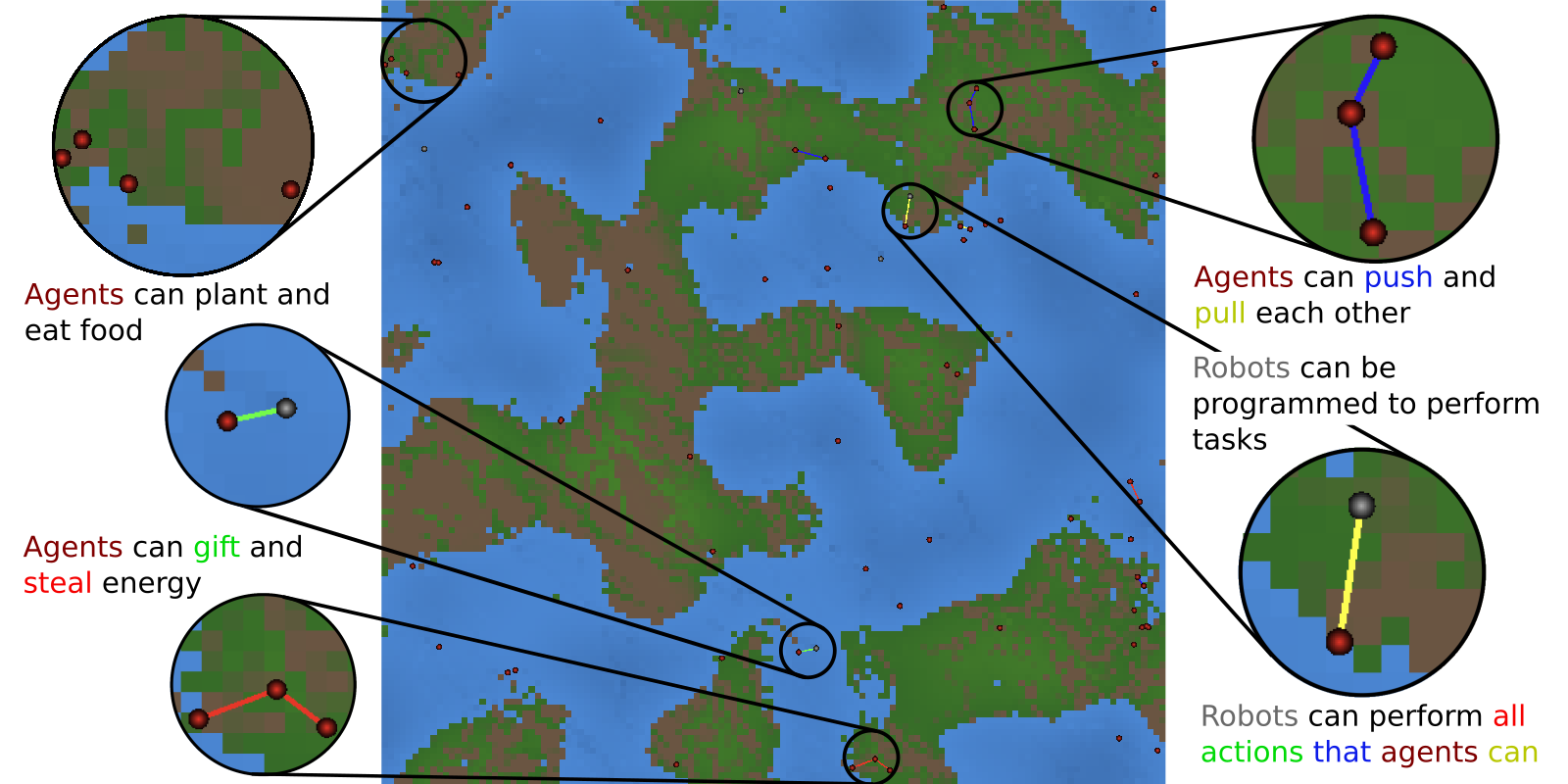}
\caption{\name is an ALife simulation containing agents that evolve through natural selection and programmable robots that can be requisitioned as tools. The color of the lines indicate the actions performed. Agents are red while bots are grey.}
\label{fig:description}
\end{figure*}

\begin{enumerate}
    \item We design and implement an agentic simulator where evolved agents must survive, and are able to interact and program robots. We show that these robots can be programmed as useful tools and express meaningful Turing-complete dynamics.
    \item We demonstrate the emergence of rudimentary agriculture, tool use, and communication. 
    \item We provide initial estimates for how important features of this simulation scale with the amount of compute provided, enabling rough estimates of expected future behaviors.
\end{enumerate}

\section{Simulation Description}

\subsection{Overview}
Our simulation consists of three primary components, terrain, agents, and robots. At every step, all agents simultaneously observe the area around them and perform a set of actions. These actions may influence other agents, the robots, or the terrain. 
Robots are programmable systems with the same action space as agents.
Agents evolve and change their behavior through the evolution of their controlling neural networks, while the terrain controls how difficult certain actions are and how much energy is available. Due to the limited amount of energy, there is selection pressure and agents that tend to eat and reproduce more will tend to pass on their genes more frequently.
Agents can also control the terrain by terraforming it, thereby changing its properties. 
Finally, robots are systems that do not evolve, but can be programmed by agents. These robots possess a large amount of potential complexity and can execute useful behaviors to help agents survive.

\subsection{Terrain}
The terrain is divided into a grid of cells, each cell possessing several attributes. The primary attributes of each cell are how much energy it has and an \textit{energy gain} amount, indicating how much energy each cell gains per timestep.
Each terrain cell also has a cost associated with each agent action. 
Finally, each cell has an associated information bit that can be read from and written to by bots but not agents.

We implement a weather and climate-like system. At every timestep, the \textit{base} terrain is slightly altered by adding a small number to the angles used to generate the Perlin noise map~\citep{perlin1985image}. The attributes of each cell also slowly regress to this \textit{base} state. The speed of this is proportional to the cell's maximum energy amount. This, for instance, can simulate long-term effects such as continental shifts.

Using this system, we can represent slowly changing landscapes, as well as the natural tendency of nature to return to its base state if not continually maintained. Finally, since the regression speed is different for different regions, the map contains high-maximum energy areas that quickly revert to their base state and lower-energy areas where changes the agents make have more permanence.

\subsection{Agents}
Agents require energy to survive and evolve through the process of natural selection.
Agents can gain energy by performing the \texttt{EAT} action on terrain cells that have energy.
Energy is consumed when performing any action (with the rate being determined by the current terrain cell).
The agents also have a continuous $(x, y)$-position, which they can control by their \texttt{MOVE\_X} and \texttt{MOVE\_Y} actions.
Agents have two types of messages, \textit{self} and \textit{other}, both of which are observed by the agent itself and other agents around it. The agent controls its own \textit{self}-message, whereas the \textit{other}-message can only be changed by other agents. If the agent chooses to send a message to other agents, it sends its self-message to the closest $N^\text{view}_\text{agents}$ agents, which updates these agents' \textit{other} message property.
This allows agents to communicate with themselves and others. As mentioned above, the agent can also spend energy to alter the properties of the terrain. Agents also age, and have to use more energy to survive as they become older.
Every timestep, agents receive an observation consisting of the terrain's attributes in a region around it, the attributes of the closest $N^\text{view}_\text{agents}$ agents and $N^\text{view}_\text{bots}$ robots, as well as its own attributes.\footnote{$N^\text{view}_\text{agents}$ and $N^\text{view}_\text{bots}$ are fixed hyperparameters of the simulation.}
\subsubsection{Network Architecture}
\begin{figure}[t]
    \centering
    \includegraphics[width=1\linewidth]{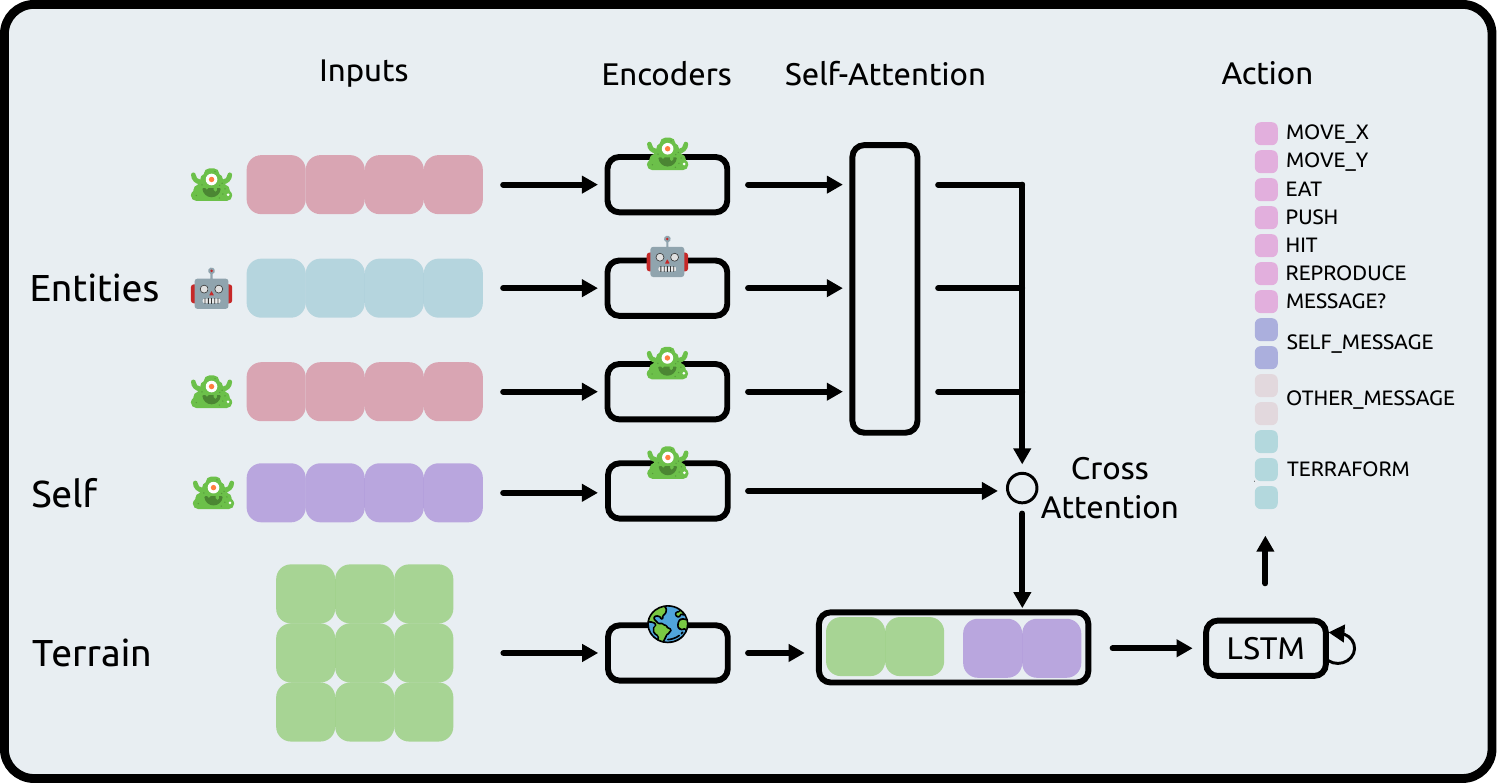}
    \caption{The agents' network architecture. Robots, agents, and the terrain each have different encoders. Entity embeddings are processed using a self-attention block, followed by cross-attention using the agent's own embedding. This, concatenated with the terrain features, is the input to an LSTM that outputs an action vector.  
    The terrain encoder is $1\times 1$ convolution followed by a fully connected layer.
    }
    \label{fig:network_arch}
\end{figure}
We parametrize agents using neural networks that process observations and output an action vector (see \cref{fig:network_arch}). All actions can be performed simultaneously, and the strength of the action's effect is determined by the magnitude of the corresponding entry. 
The network architecture consists of different encoders for agents, robots, and the terrain. The entity embeddings are processed by a self-attention and multi-headed attention block~\citep{vaswani2017Attention}. The result is concatenated with the terrain features and passed to an LSTM~\citep{hochreiter1997long} to incorporate memory.
While the agent's networks do not change during their lifetimes, due to the use of a recurrent network, they are able to adapt their behavior when encountering the same observation multiple times.

\subsubsection{Reproduction}
Since crossover for fixed-topology neural networks is challenging~\citep{Haflidason2009On,pretorius2024neural}, our agents reproduce asexually, and must have a minimum amount of energy before they can execute the \texttt{REPRODUCE} action. 
Reproduction copies the agent's weights to a child agent, and random perturbations are also added to these weights. 
We reinitialize the population---with random networks---if all agents die.

\subsection{Robots}
We design the robots to have a similar action space to the agents; however, since robots do not reproduce, and we wish them to be programmable, how they obtain their actions must be different to the agents. To simulate the technological advantage of machines, robots update multiple times for every agent update cycle and do not use energy.
We decide upon the following scheme and showcase its theoretical complexity and practical uses in the next section.

Each robot has a \textit{program} and \textit{memory}, each of the same size $N_\text{prog}$. At every step, each robot receives messages from the two closest entities to it, breaking ties using the x-position. This message is the \texttt{SELF\_MESSAGE} of an agent or the \textit{memory} of a bot; each of these is also of size $N_\text{prog}$.

The program is a description of a function $f: \mathbb{R}^{N_\text{prog}} \times \mathbb{R}^{N_\text{prog}} \times \to \mathbb{R}^{N_\text{prog}}$, where the action $a = f(\text{mem}, m_1, m_2)$ is the action performed. This action, if the \texttt{WRITE\_SELF\_MESSAGE} entry is set, can also update the robot's memory. Whenever another robot or agent sends this robot a message, it is interpreted as changing the robot's program.

We note here that the dimension of these messages $N_\text{prog}$ is always more than the action's dimensionality $N_\text{act}$, therefore, only the first part of the output is used as the action. 
The final entry, in particular, is interpreted as solely an information bit that allows robots to store and manipulate information.

\subsubsection{Instructions}

We have the following instructions:
\begin{itemize}
    \item \texttt{COPY}: $f(\text{mem}, m_1, m_2) = m_1$
    \item \texttt{NOOP}: $f(\text{mem}, m_1, m_2) = \text{mem}$
    \item \texttt{PRODUCT}: $f(\text{mem}, m_1, m_2) = \text{mem} \odot m_1$
    \item \texttt{FMA}: $f(\text{mem}, m_1, m_2) = \text{mem} \odot m_1 + m_2$
    \item \texttt{XOR}: $f(\text{mem}, m_1, m_2) = (2(\text{mem}^i \oplus m_1^i) - 1) \odot m_1 \odot \text{mem} + (1 - m_1) \text{mem}$, where $\oplus$ is logical XOR, and the superscript $i$ indicates the information bit of the message.
    \item \texttt{NAND}: $f(\text{mem}, m_1, m_2) = 1 - (m_1 \odot m_2)$
\end{itemize}
The final operation, \texttt{LOOKUP}, uses the information bit from $\text{mem}$, $m_1$ and $m_2$ to construct a 3-bit number, from $0$ to $7$ inclusive, and uses this index to look up into a table as specified by the program. The result is written to the information bit of the action and replaces the bot's current memory if the \texttt{WRITE\_SELF\_MESSAGE} entry of the action vector is set. This allows bots to compute complex functions, and to store the result in their information bit.

\section{Analyzing Environment Complexity}
Here we discuss the capabilities of the robots in \name.  We begin by illustrating useful and practical bots.  We then go on to prove our simulation is Turing-complete~\citep{turing1936Computable}, by showing that it can execute Rule 110 ~\citep{cook2004Universality,cook2009Concrete}. Finally, we describe how robots can also compute arbitrary boolean functions. 

\subsection{Useful Machines}\label{sec:useful_machines}
\begin{figure}
    \centering
    \begin{subfigure}{0.32\linewidth}
        \includegraphics[width=1\linewidth]{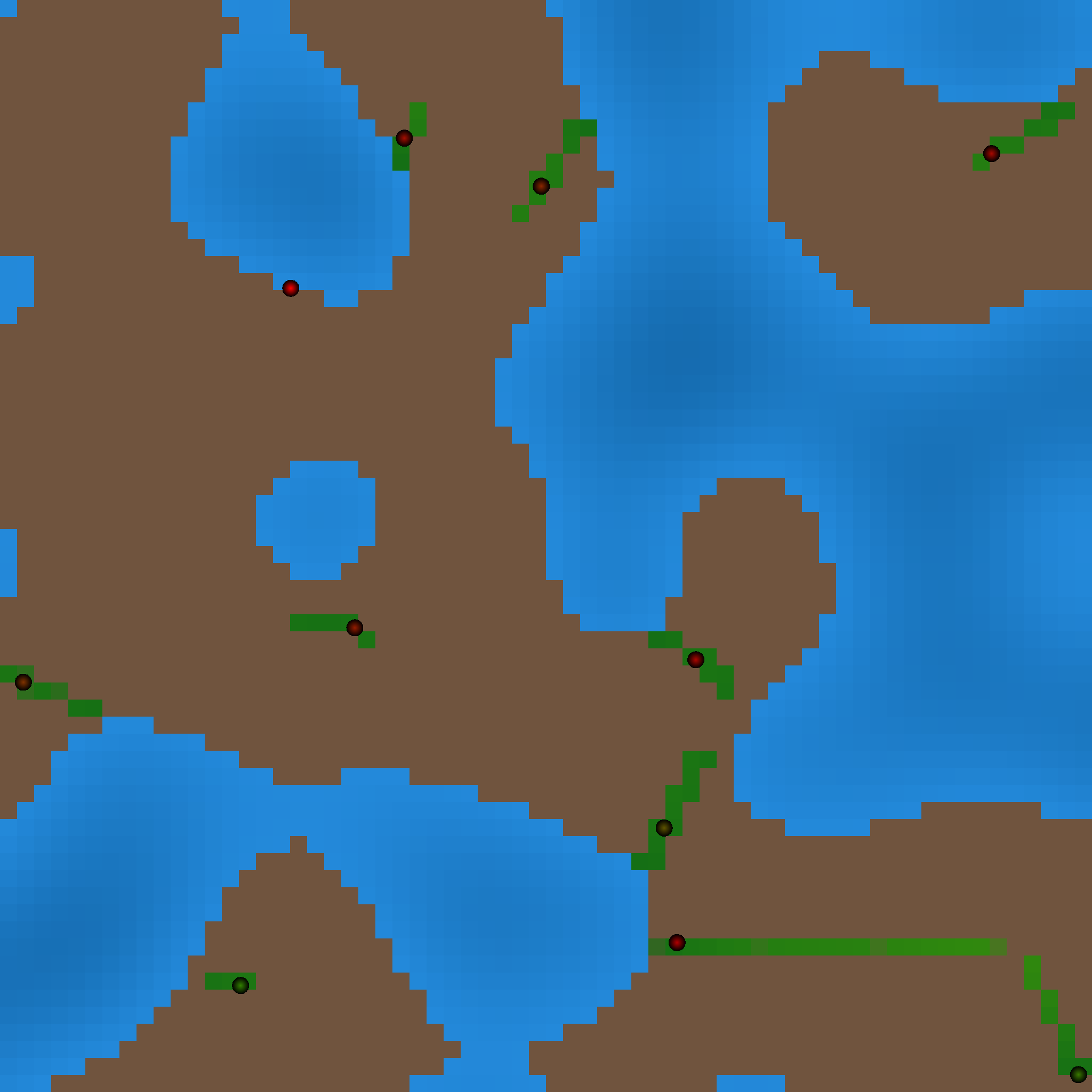}
        \caption{}
    \end{subfigure} 
    \begin{subfigure}{0.32\linewidth}
        \includegraphics[width=1\linewidth]{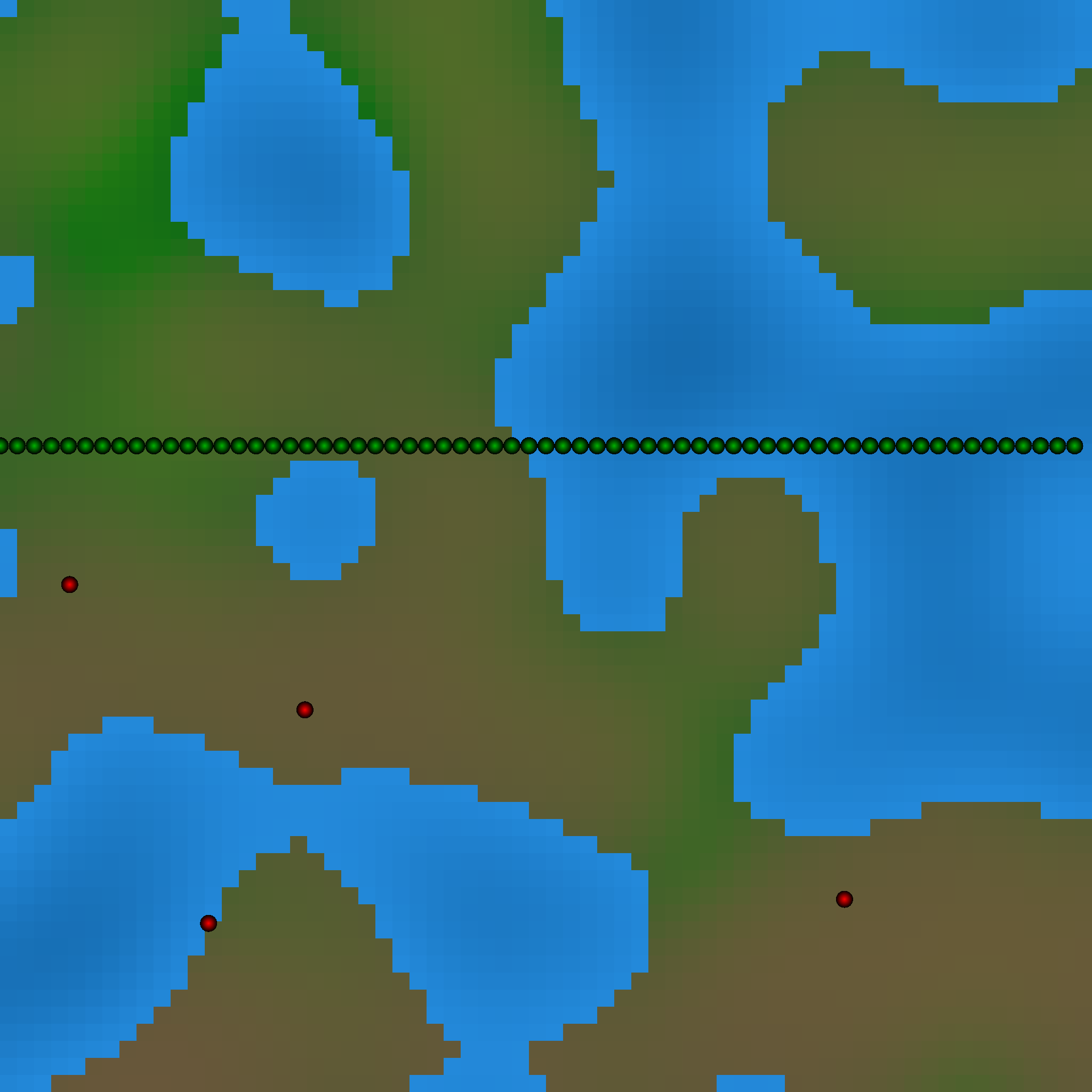}
        \caption{}
    \end{subfigure} 
    \begin{subfigure}{0.32\linewidth}
        \includegraphics[width=1\linewidth]{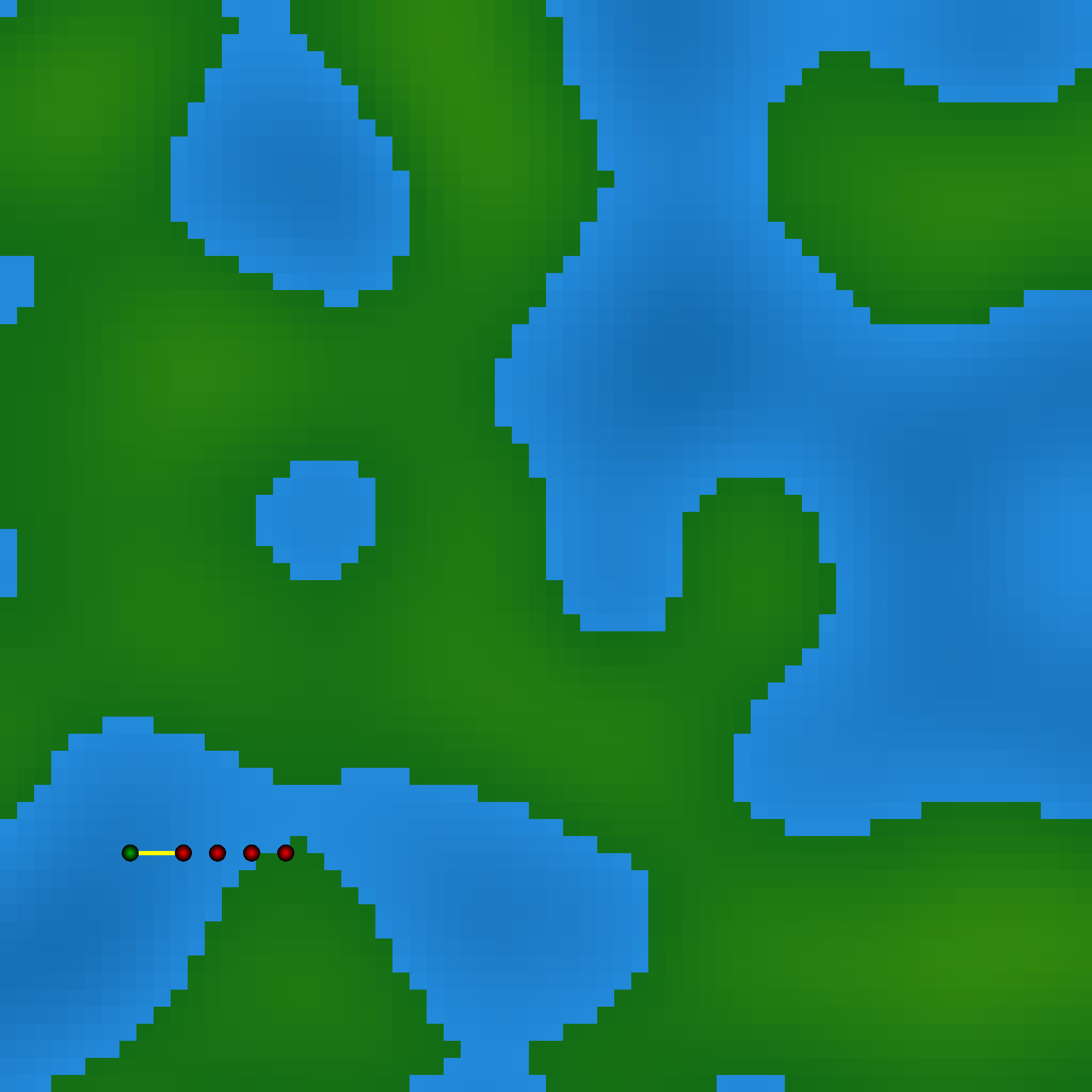}   
        \caption{}
    \end{subfigure}
    
    \caption{Three snapshots of manually-designed useful bots. (a) shows patrolling bots, (b) shows mass terraforming and (c) shows the transport bot, which is pushing the agents.}
    \label{fig:useful}
\end{figure}

In practice, we expect the programs created by agents to be relatively simple at first. 
However, to illustrate what is practically possible, we manually design some useful robots that can easily be constructed, see \cref{fig:useful}.

\subsubsection{Automated Terraforming}
These robots are programmed to terraform a large region of the map. This can be implemented by programming robots to move in a consistent direction (e.g. down the map), and perform the  \texttt{TERRAIN\_ENERGY\_GAIN} action. This leads to the terrain becoming more fertile--- leading to more food over time.

\subsubsection{Patrolling \& Oscillating}
Robots can also execute more complicated oscillatory behavior, by using the terrain's information bits as waypoints to oscillate over an arbitrary line. This is achieved by reading the current terrain bit, and using the \texttt{XOR} instruction, which inverts the action when paired with an appropriate $m_1$ sent by the closest agent.

\subsubsection{Transportation}
Robots can also be used to transport agents in a more energy-efficient manner than walking. Suppose the robot's memory is zero, except for the entries associated with \texttt{MOVE\_X} and \texttt{PUSH}. The robot's program is the one defined as $f(\text{mem}, m_1, m_2) = \text{mem} \odot m_1 + m_2$. This means that the robot will push and move nearby agents whenever they send messages with the move and push entries being nonempty.

\subsubsection{Communicating}
Robots can also propagate information across space. We implement this proof-of-concept by using the lookup table instruction, with the lookup table simply copying the information bit of $m_1$. Arranging the robots in a chain such that the robot to the left of it is closer than the one to the right allows the information bit to pass from the left to the right across the map.

\subsection{Turing-Complete Computation}\label{sec:turing_complete}
We now move on to proving that \name can facilitate universal computation by reducing it to Rule 110---a common technique that has been used in several prior settings; for instance, in Baba is you~\citep{rodriguez2019BABA,su2023New}, the Micron Automata Processor~\citep{wang2015Cellular} and Petri Nets~\citep{zaitsev2018Simulating}.

\subsubsection{Constructing Rule 110}
Elementary cellular automata~\citep[ECA]{wolfram2002Anew} are simple, one-dimensional rules that can lead to complex patterns. An ECA is defined on a grid, with an initial state in the top row, and a local transition rule that transforms the row into the next one; every time step this is applied and a new row is appended. In ECAs, every cell is binary-valued, and the local transition rule of a particular cell depends on it, as well as its left and right neighbors. Using these three binary digits $a$, $b$ and $c$, we can construct a binary number ${abc}_2$ such that $0 = {000}_2 \leq {abc}_2 \leq {111}_2 = 7$. Each ECA is then an 8-dimensional lookup table, giving the next value of the center cell for each possible three-digit binary number. Rule 110 has the pattern described in \cref{tab:rule110}:
\begin{table}[h]
    \centering
    \caption{Rule 110}
    \label{tab:rule110}
    \resizebox{\linewidth}{!}{\begin{tabular}{lccccccccc}
    \toprule
         Pattern (lcr) & 111 & 110 & 101 & 100 & 011 & 010 & 001 & 000  \\
     \midrule
        Next Value & 0 &	1&	1&	0&	1&	1&	1&	0\\
     \bottomrule
    \end{tabular}}
\end{table}

\citet{cook2004Universality} proved that Rule 110 is capable of universal computation, i.e., that it is Turing complete.
Here we describe a way in which a particular configuration of \name results in an implementation of Rule 110---showing that it, too, is capable of universal computation. The core idea is to use the terrain's information bit to store the previously computed rows, and to have a single row of robots act as the latest state while moving down the map.

\noindent \textbf{Program}
The program has the \texttt{LOOKUP\_TABLE} instruction bit set as $1$, all other instruction bits set to zero, and the lookup table is defined as in \cref{tab:rule110}.

\noindent \textbf{Memory}
The memory is empty, except for three locations:
\begin{enumerate}
    \item $\texttt{MOVE\_Y} = 1$: In order to move in the y-direction.
    \item $\texttt{WRITE\_TERRAIN} = -1$: To write the previously-computed bit to the terrain (note that $+1$ indicates reading and $-1$ indicates writing).
    \item $\texttt{UPDATE\_MEMORY} = 1$: To save the updated cell state to the robot's memory.
\end{enumerate}

\noindent \textbf{Result}
The result of this construction is that at each step, each bot performs the following in order: \begin{enumerate}
    \item Compute Rule 110 using the information bits of the robot immediately to its left and right, as well as its own bit. 
    \item Write the updated state to its memory's information entry.
    \item Write this information bit to the terrain
    \item Move one position forward in the x-direction
\end{enumerate}

\begin{figure}
    \centering
    \begin{subfigure}{0.32\linewidth}
        \includegraphics[width=1\linewidth]{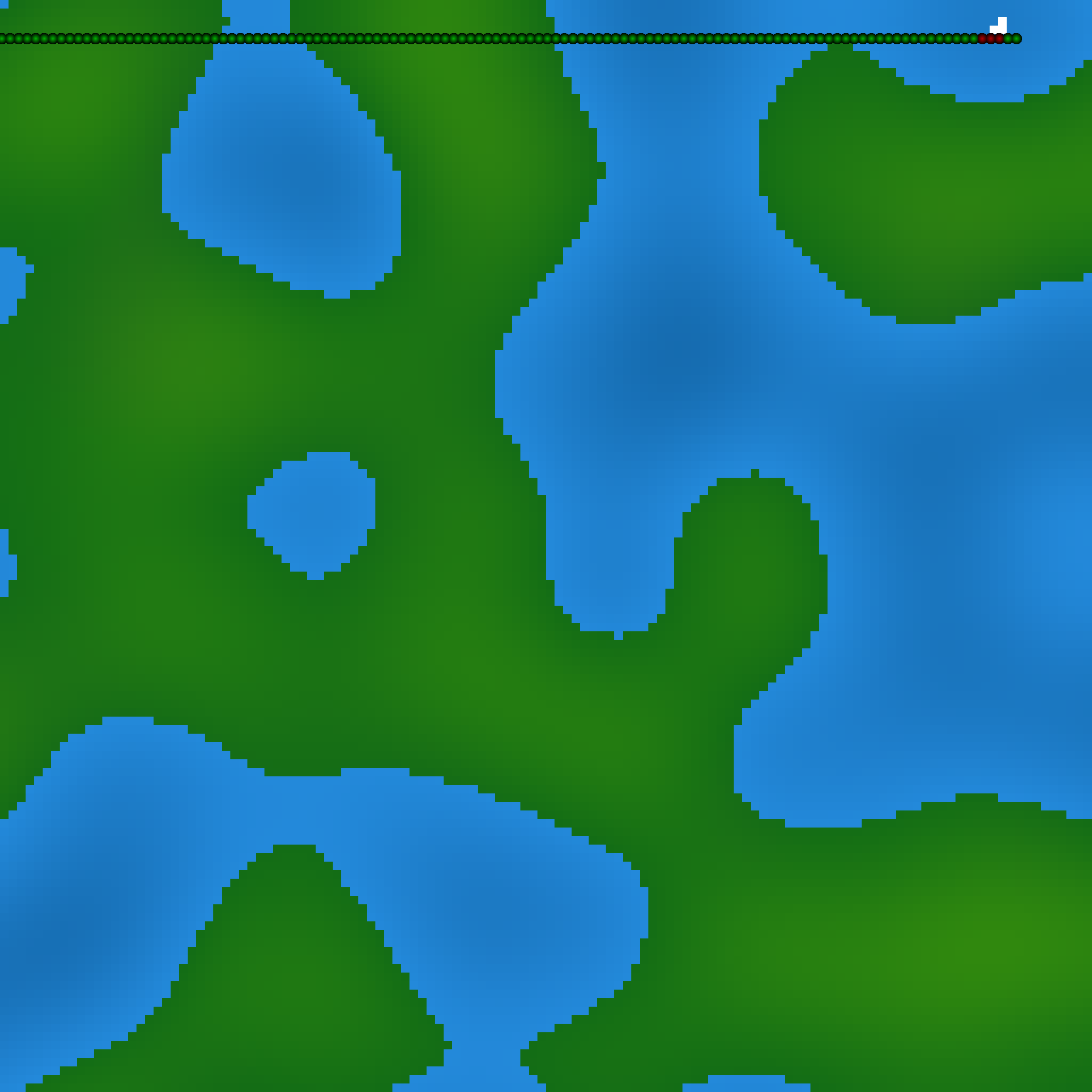}
        \caption{}
    \end{subfigure} 
    \begin{subfigure}{0.32\linewidth}
        \includegraphics[width=1\linewidth]{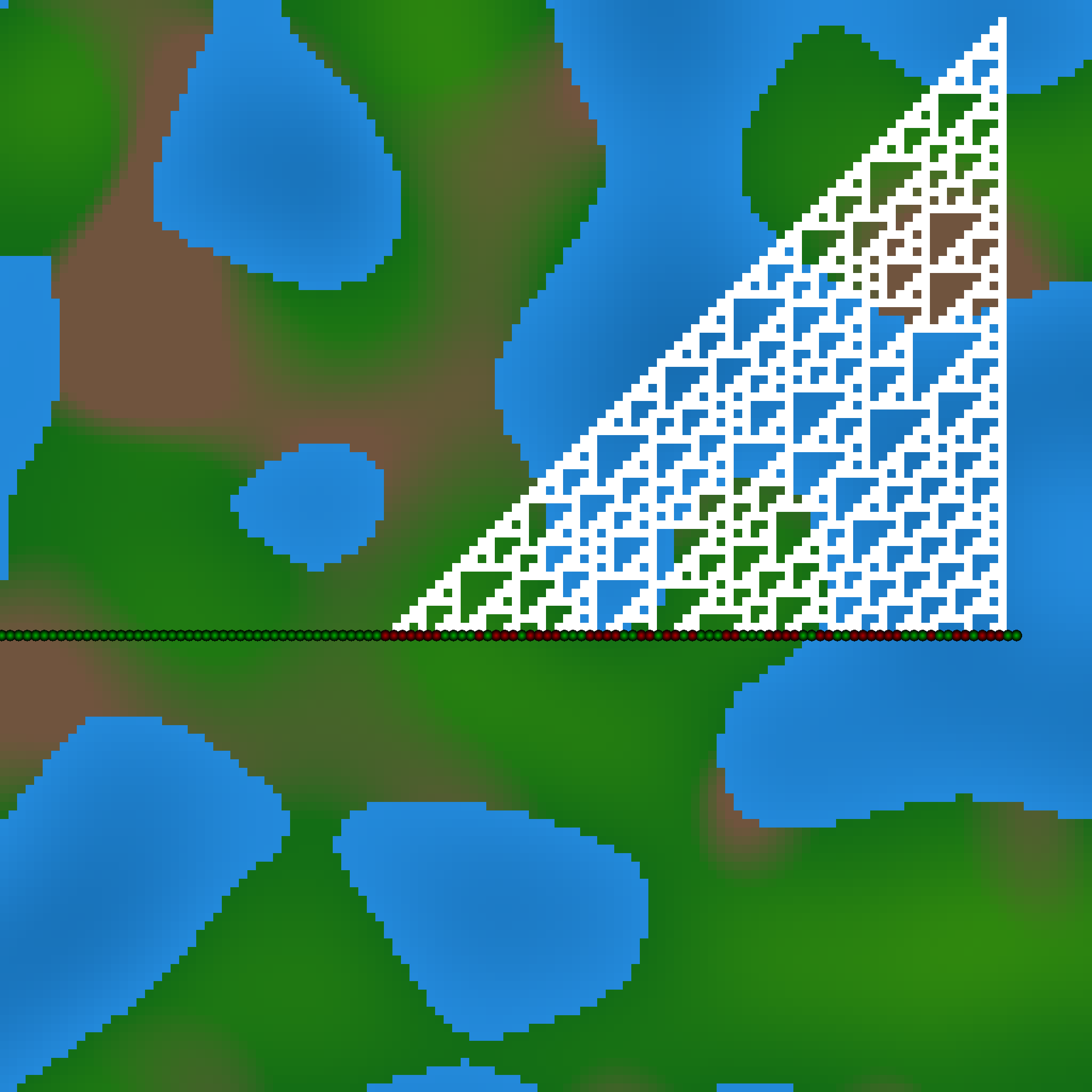}
        \caption{}
    \end{subfigure} 
    \begin{subfigure}{0.32\linewidth}
        \includegraphics[width=1\linewidth]{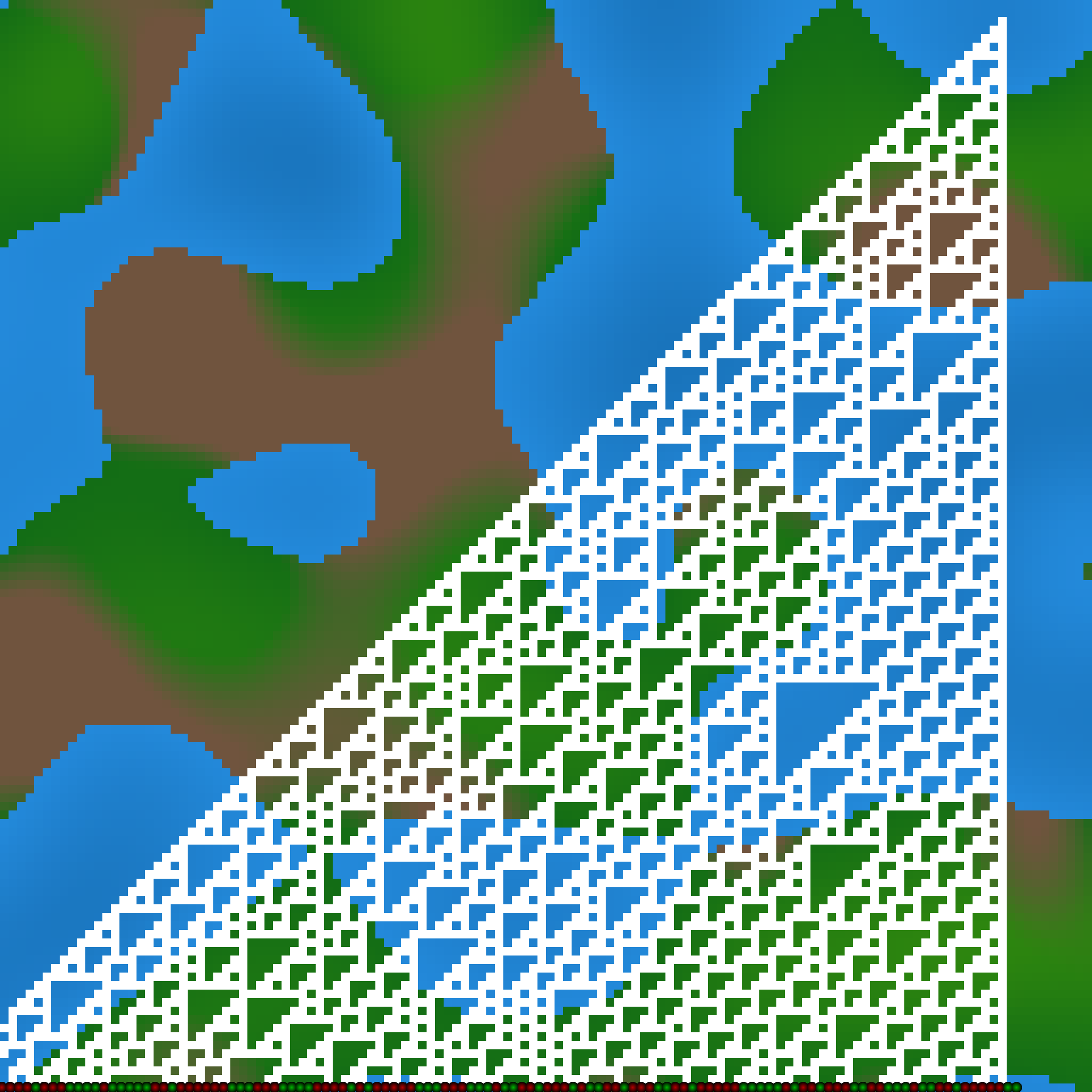}
        \caption{}
    \end{subfigure}
    \caption{Three snapshots of the computation of Rule 110. In this figure, terrain bits are explicitly rendered.}
    \label{fig:rule110}
\end{figure}

Therefore, the robots collectively implement Rule 110, with the previous iterations' results being saved on the terrain bits of the world. The initial information bit in each robot's memory defines the initial state of the simulation, and can be set arbitrarily.
Under the assumption of an infinitely large simulation, this system of bots is Turing-complete, due to it being able to implement Rule 110 with an arbitrary initial pattern~\citep{cook2004Universality}. An illustration of this process is presented in \cref{fig:rule110}.

\subsection{Functional Completeness}\label{sec:functional_complete}
Functional completeness of a set of logical operators states that all possible boolean functions can be realized by composing elements from this. Functionally complete of complete operators include $\{ \texttt{NAND} \}$, $\{ \texttt{NOR} \}$, $\{ \texttt{OR}, \texttt{AND}, \texttt{NOT} \}$~\citep{enderton2001Mathematical}.

\subsubsection{NAND} 
We note that one of our instructions described above---the lookup table---can implement any three-bit truth table, which includes \texttt{NAND} between two inputs, which we refer to as the \texttt{NAND} instruction.
We now describe a way to compose logical operators, resulting in functional completeness. 

\subsubsection{Composition}
Suppose we have a vertical line of $n$ input bots $A_1, \dots, A_n$. We can compute any possible function by having at most $n$ columns of $n$ bots each, that process these input bits, such that the final column has one bot which represents the output. This final bot could then write to the terrain.
We describe now how to perform the two necessary operations, \texttt{NAND} and passthrough (i.e., identity).

Consider two robots $A$ and $B$ in row $i$ that contain the input bits in their information slot. Suppose robot $R^{i+1}_j$---in row $i+1$---is positioned such that its closest two agents are $A$ and $B$. Then, by using the \texttt{NAND} program, it can compute $A$ \texttt{NAND} $B$ and store the result in its memory.

To pass bit $A$ unchanged to the next row, suppose that bot $A$ is closest to bot $R^{i+1}_j$; it can then use the \texttt{COPY} action to copy the bit unchanged (this could also be implemented using the lookup table to copy the closest robot's bit).

\subsubsection{Temporal Order}
Due to the temporal nature of our simulation, we cannot compute an arbitrary function in one step. We note that in our above construction, initially, only the input column is correct. Every step ensures that the subsequent column computes the correct function. At the end, the final bot's memory would be correct.
Using these two operations, and the temporal structure, the robots in \name can compute any $n$-variable binary-valued function for arbitrary $n$.

\begin{figure*}
    \centering
    \includegraphics[width=1\linewidth]{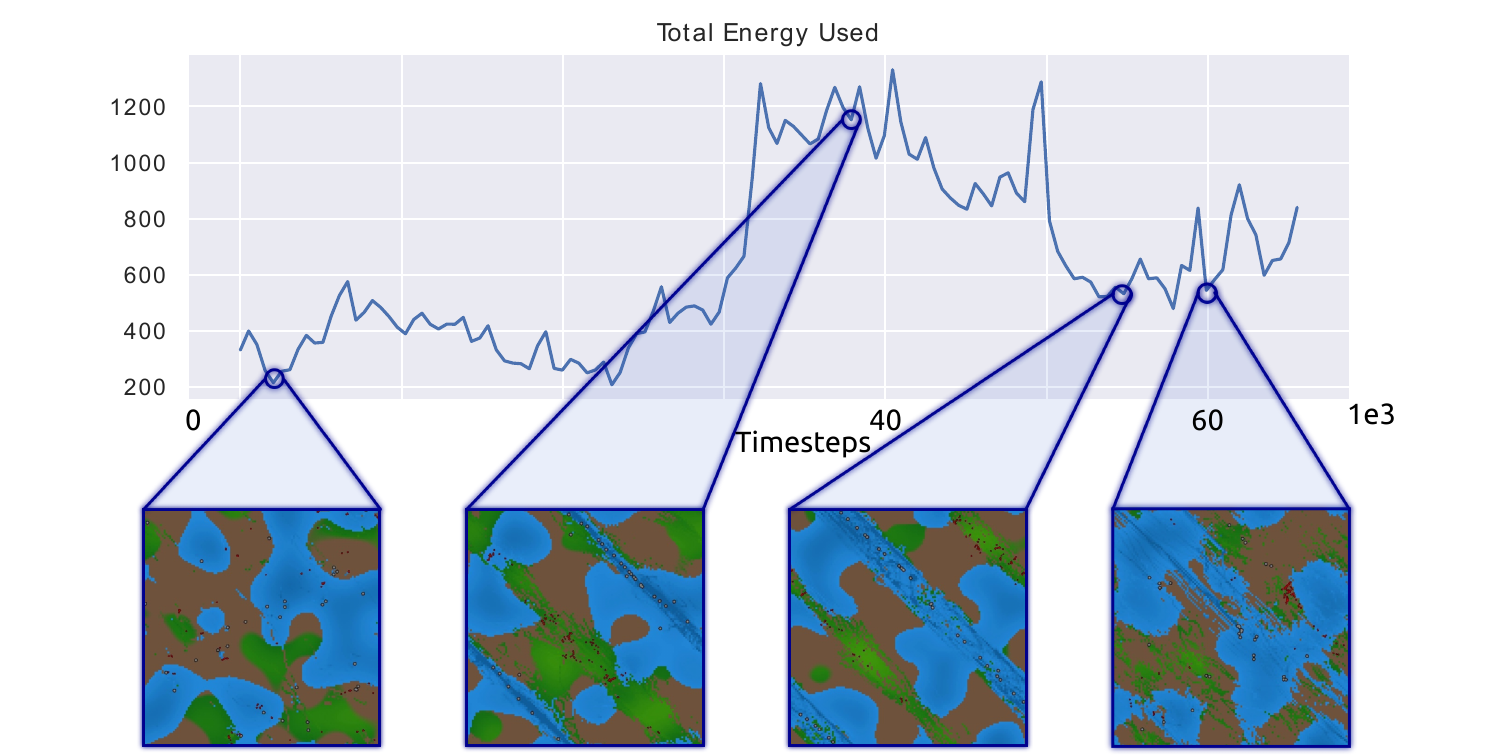}
    \caption{Snapshots of the simulation at various points in time. At first, the world state is effectively random. At some point, however, agents start forming a diagonal bridge, and travel along it. This coincides with a large spike in energy usage. Later on the bridge collapses and the total energy usage declines.}
    \label{fig:qual-figure}
\end{figure*}

\section{Results}
In this section, we present the empirical results obtained when running \name.
We run the simulation for $2^{16}$ timesteps using a single NVIDIA A100 GPU. We first describe behaviors that we observed, with 128 agents and 32 bots, and then examine quantitative metrics of the complexity of the simulation. Finally, we observe how these results change with scaling the number of agents in the environment.

\subsection{Emergent Behaviors}

\subsubsection{Agriculture and Terraforming:} \cref{fig:qual-figure} shows that the agents manage to substantially modify their environment, creating structured regions in which food is produced. We find that this corresponds to a period in which the agents construct striped diagonal regions over which they travel. 

\subsubsection{Tool-Use:} 
\begin{figure}[h]
    \centering
    \includegraphics[width=0.85\linewidth]{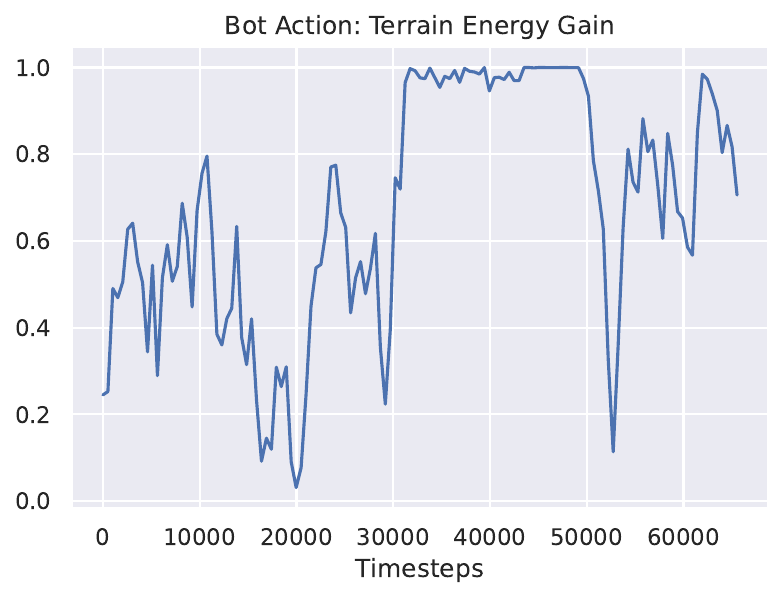}
    \caption{The average value of all of the bots' ``Terrain Energy Gain'' action, which increases the amount of energy a given position produces. We observe that the time period in which this is high corresponds to a peaked region in \cref{fig:qual-figure} where the bots have constructed a bridge.}
    \label{fig:bot_actions}
\end{figure}

In \cref{fig:bot_actions} we visualize the behavior of the programmed bots. In particular, we find that the bots play a significant role in the stable terraforming behavior we found in \cref{fig:qual-figure}, as they move at the same diagonal stripes as the agents while also helping with food production.

\subsubsection{Communication} 
In \cref{fig:main_results:commsal} we use ``saliency'' (i.e., the derivative of the agent's action with respect to the input \citep{simonyan2013deep}) as a measure of the use of communication. We find that, over time, agent behaviors tend to be more sensitive to their communication channels, suggesting that they can use the channel to influence each other.

\subsection{Scaling Results}

\begin{figure*}
    \centering
    \begin{subfigure}{0.24\textwidth}
        \includegraphics[width=1\linewidth]{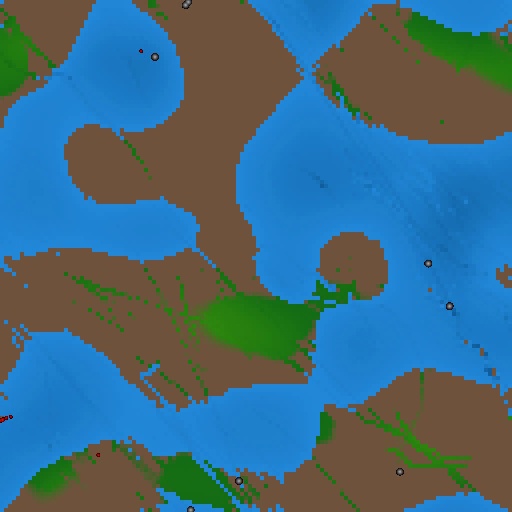}
        \caption{32 Agents}
    \end{subfigure} 
    \begin{subfigure}{0.24\textwidth}
        \includegraphics[width=1\linewidth]{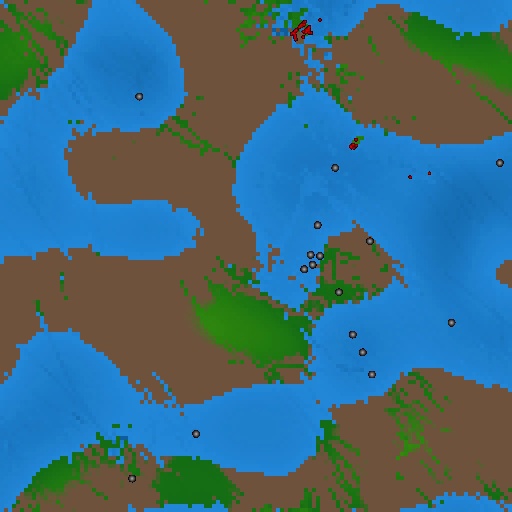}
        \caption{64 Agents}
    \end{subfigure} 
    \begin{subfigure}{0.24\textwidth}
        \includegraphics[width=1\linewidth]{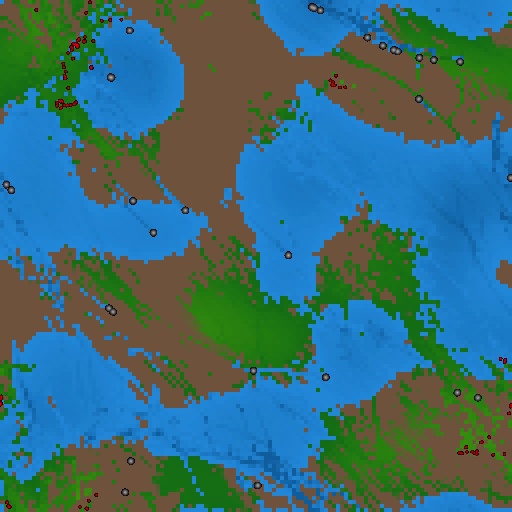}
        \caption{128 Agents}
    \end{subfigure}
    \begin{subfigure}{0.24\textwidth}
        \includegraphics[width=1\linewidth]{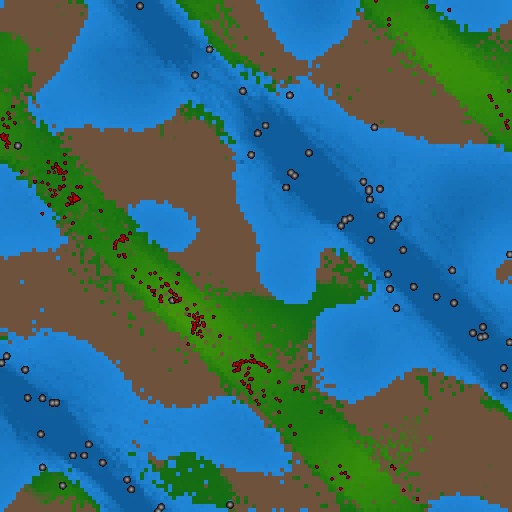}
        \caption{256 Agents}
    \end{subfigure}

    \caption{Visualization of the results of scaling the number of agents after 18000 steps. At too few agents, we do not observe any meaningful emergent patterns. However, as we increase the number of agents, large-scale patterns emerge, such as the diagonal stripes in (d).}
    \label{fig:scaling-metrics}
\end{figure*}

\begin{figure*}[h]
    \centering
    \begin{subfigure}{0.32\linewidth}
        \includegraphics[width=1\linewidth]{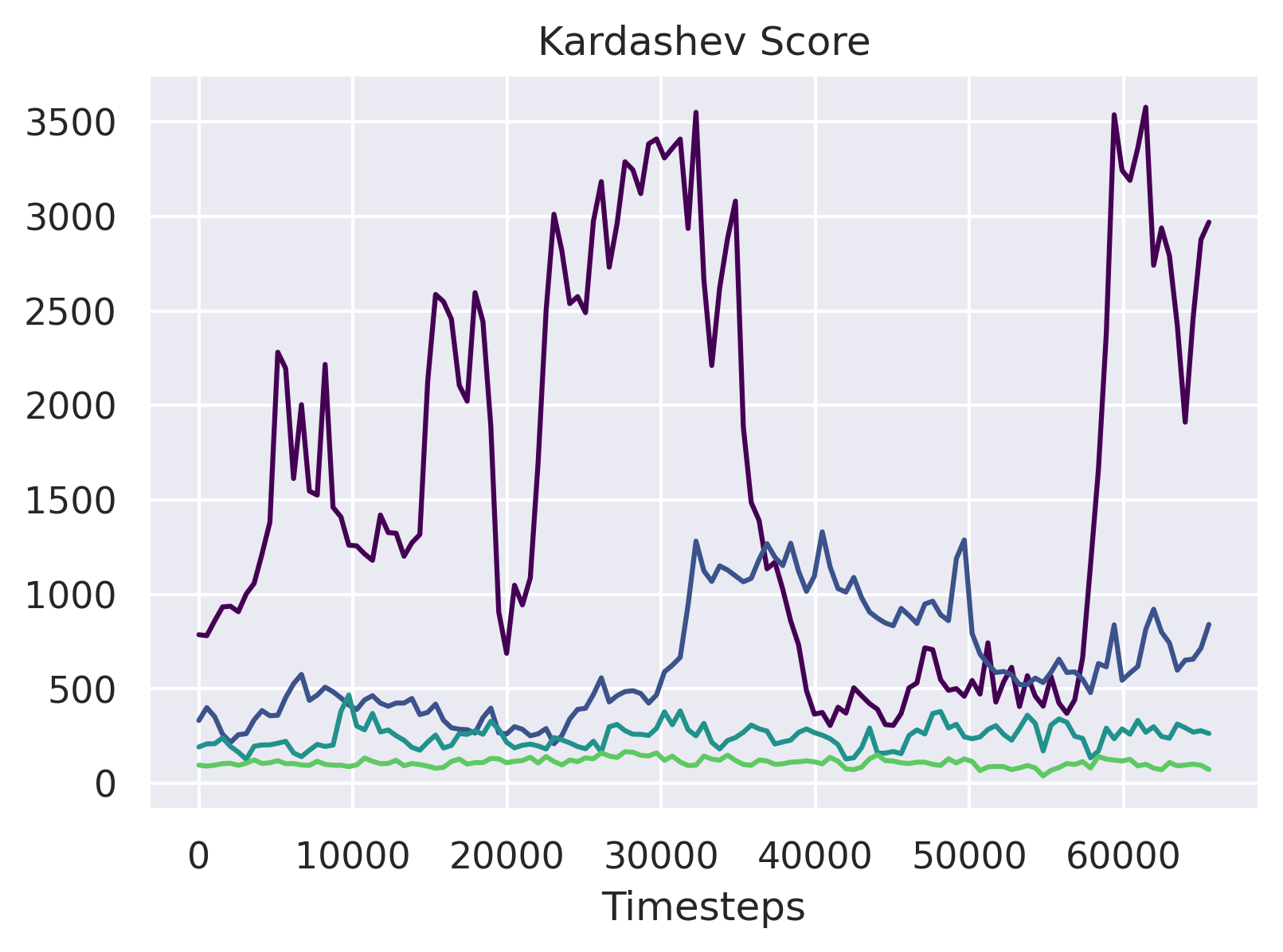}
        \caption{}
        \label{fig:main_results:kard}
    \end{subfigure}
    \begin{subfigure}{0.32\linewidth}
        \includegraphics[width=1\linewidth]{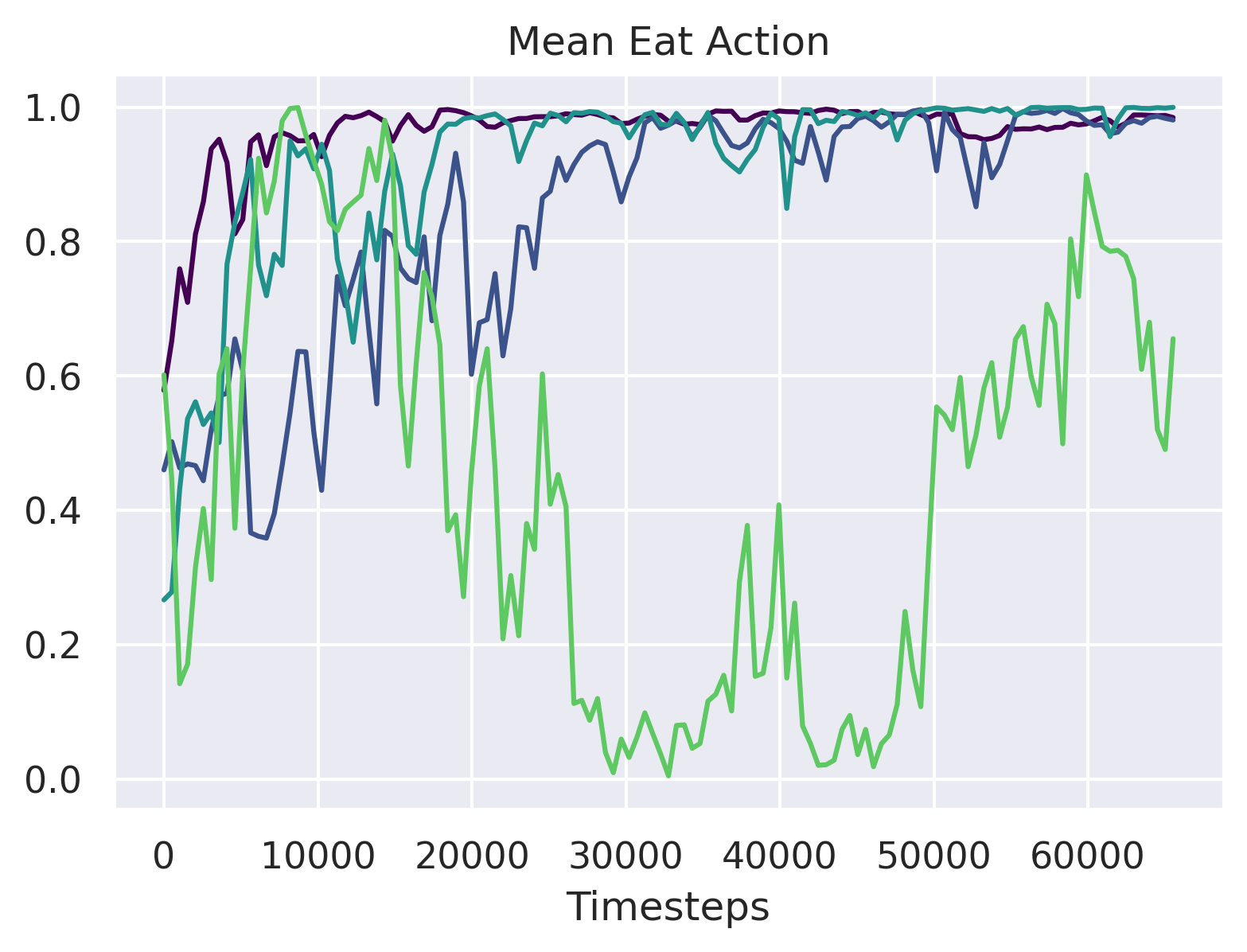}
        \caption{}
        \label{fig:main_results:eat}
    \end{subfigure}
    \begin{subfigure}{0.32\linewidth}
        \includegraphics[width=1\linewidth]{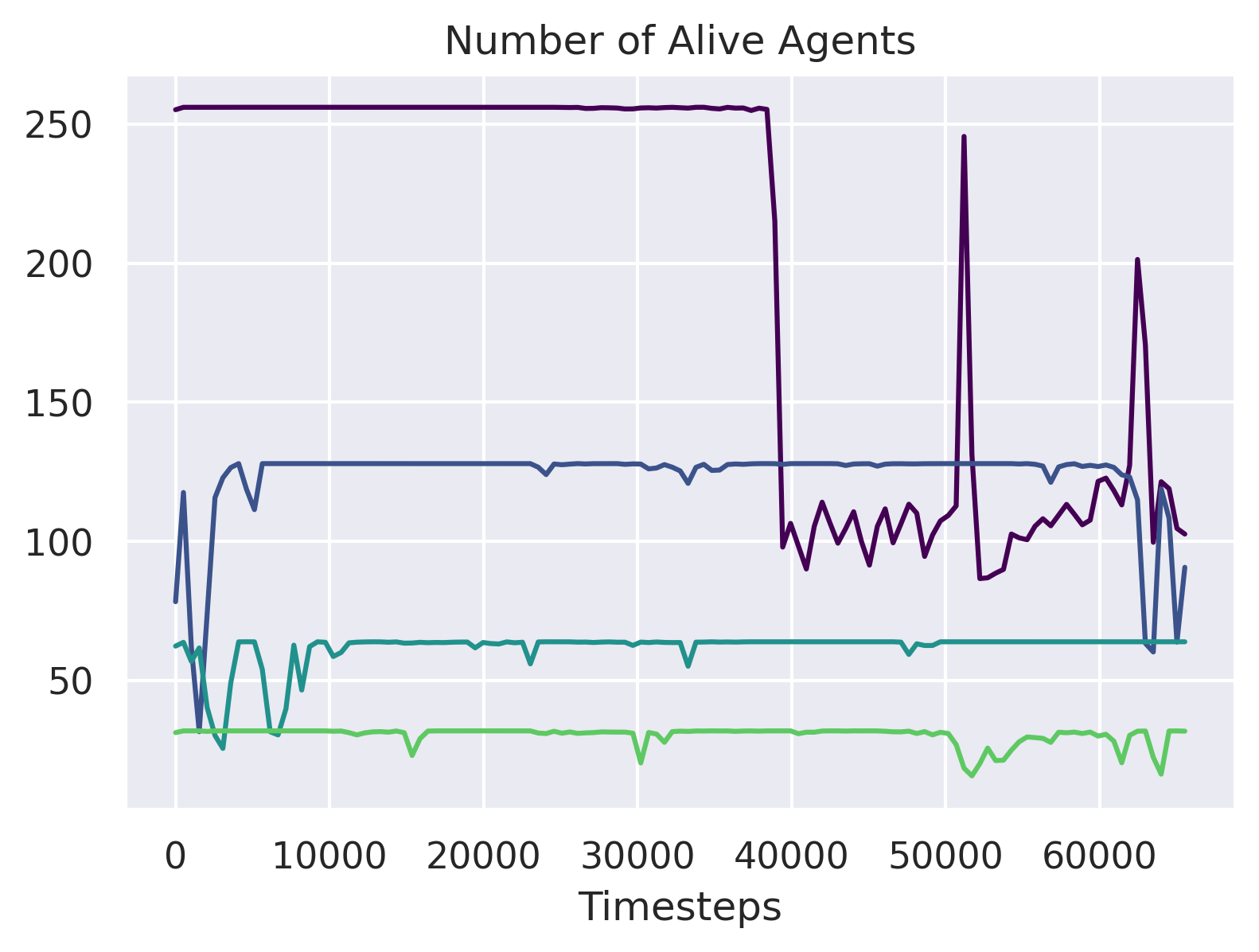}
        \caption{}
        \label{fig:main_results:num_agents_mean}
    \end{subfigure}

    \begin{subfigure}{0.32\linewidth}
        \includegraphics[width=1\linewidth]{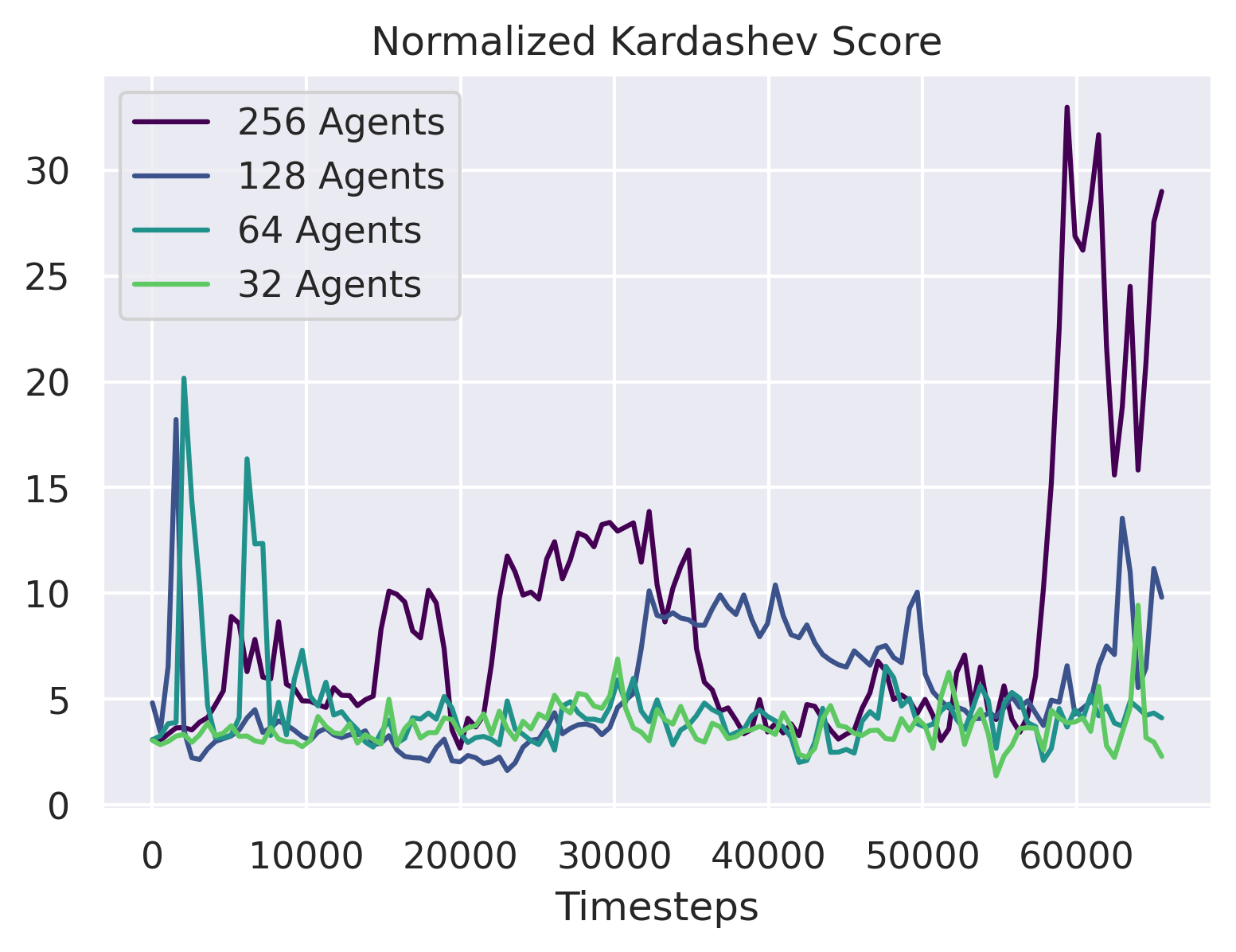}
        \caption{}
        \label{fig:main_results:kardnorm}
    \end{subfigure}
    \begin{subfigure}{0.32\linewidth}
        \includegraphics[width=1\linewidth]{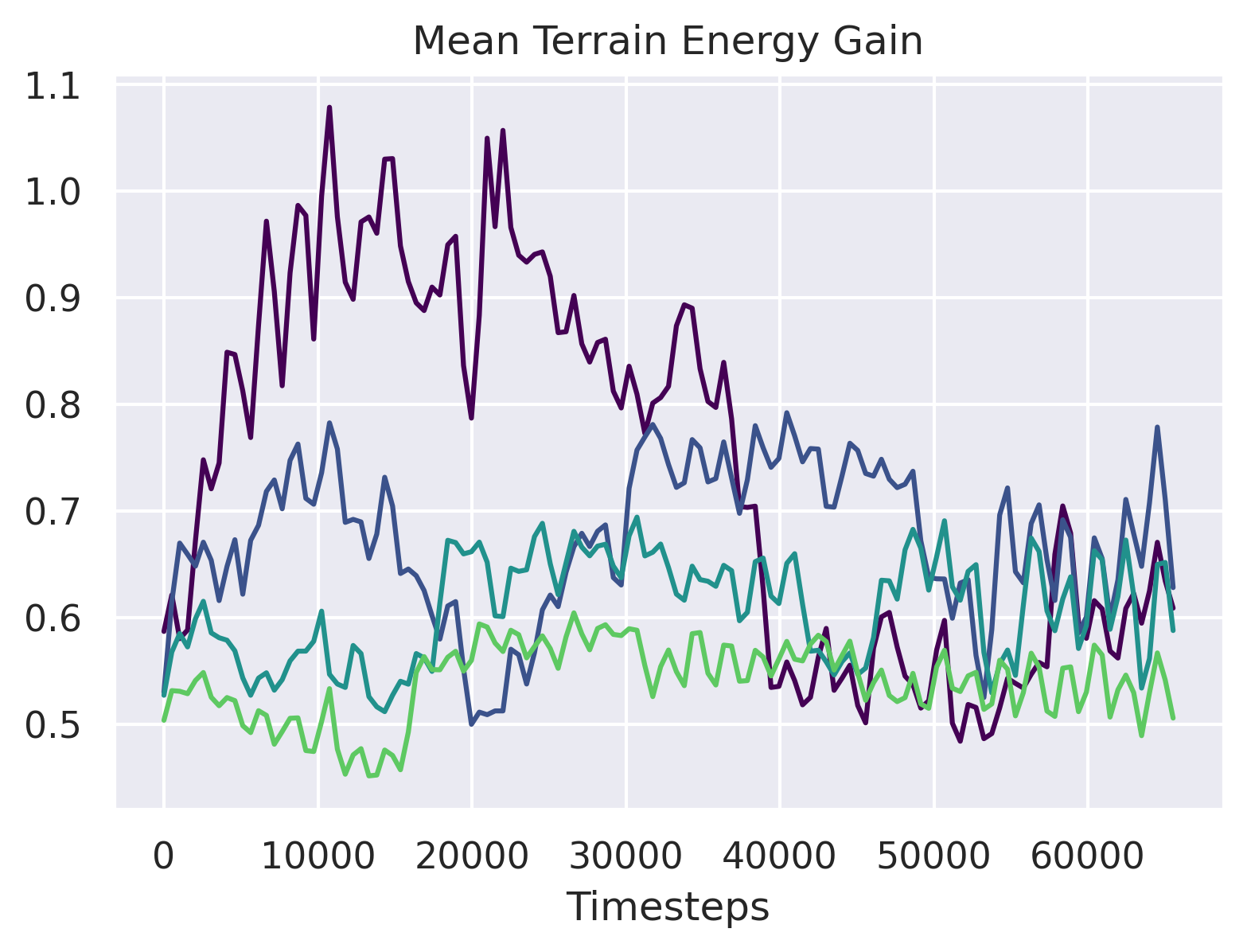}
        \caption{}
        \label{fig:main_results:terrain}
    \end{subfigure}
    \begin{subfigure}{0.32\linewidth}
        \includegraphics[width=1\linewidth]{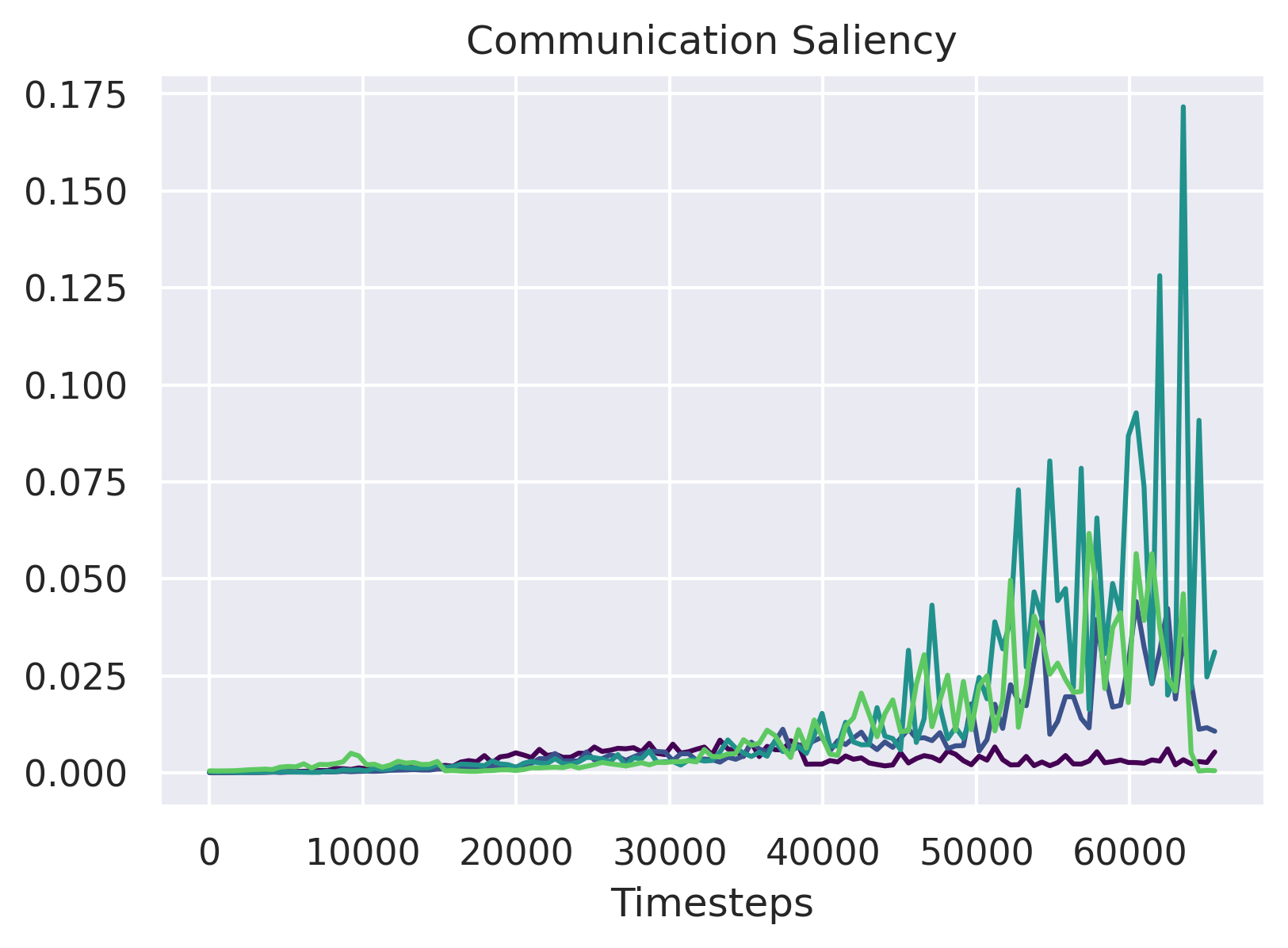}
        \caption{}
        \label{fig:main_results:commsal}
    \end{subfigure}

    \caption{Quantitative metrics over time for different numbers of agents. Kardashev score in (a) is a measure of the total amount of energy used; this is normalized by the number of alive agents in (d).}
    \label{fig:main_results}
\end{figure*}

To quantitatively measure the complexity of the simulation, we consider several metrics in Figure \ref{fig:main_results}.
\begin{enumerate}
    \item \textit{Energy Consumption}: The total amount of energy available to a civilization is often considered a marker of technological advancement~\citep{kardashev1964Transmission,gray2020Extended}. Inspired by this, we measure the amount of energy per agent consumed by the population of \name.
    \item \textit{Terrain}: Since agents have the ability to transform the terrain, we measure how much the terrain changes over time, focusing on the total amount of energy produced.
    \item \textit{Action Selection}: We analyze the distribution of agent actions, i.e., how agents are spending their energy and how this changes over time. More specifically, we show how often agents are using the EAT action, as changes in this value would be evidence for selection.
    \item \textit{Input Saliency} How much the agents use the messages received from other agents in deciding their actions, computed as the partial derivative of the action vector with respect to the observations. In particular, we investigate \textit{communication} saliency (e.g. the impact of the messages on the action) and \textit{bot} observation saliency.
\end{enumerate}

We present our primary results in \cref{fig:main_results}, indicating the value of each metric across time. We also vary the number of agents to provide an indication of how the results vary with the amount of compute we provide.
First, in \cref{fig:main_results:kard,fig:main_results:kardnorm}, we see that the amount of energy used (both overall and per agent) remains relatively consistent if we have less than 256 agents. 
If we have 256, there are more pronounced peaks and troughs, indicating that the agents go through phases of using a large amount of energy, followed by sharp declines. We find this is caused by the primary food source becoming exhausted.

In most cases, except if there are only 32 agents, the agents quickly learn to perform the eat action almost constantly (\cref{fig:main_results:eat}). We find that the terrain's average energy gain amount---indicating how fertile the land is---- has a decreasing trend with 256 agents, but remains relatively constant if we have fewer agents.

In \cref{fig:main_results:num_agents_mean}, we further find that most settings result in the maximum number of agents being alive, but that there are some spikes, and a notable drop in the number of alive agents, coinciding with the large decrease in total energy used. Notably, this shows that the population does not entirely die out, and the automatic reinitialization does not occur.

Finally, we note that communication saliency (\cref{fig:main_results:commsal}) increases gradually over the course of the simulation. Interestingly, with 256 agents, the communication saliency remains relatively constant at near zero, possibly indicating that the agents have learned to not use communication if there are too many potentially random agents.

In summary, we note that there are differences in behavior and metrics as we increase the number of agents in the simulation. We further see non-trivial progression over time, suggesting that running the simulation for longer and with more agents would be interesting.

\section{Related Work}

Early instantiations of the ALife concept include Tierra~\citep{ray1996approach} and Avida~\citep{ofria2004avida}, which evolved computer programs that compete for finite resources.  Geb~\citep{channon2006unbounded} was the first simulation said to pass the activity statistics test \citep{bedau11998classification} and to be a truly open-ended system.  However, all these simulations are relatively simple.
Despite their simplicity, however, these simulations can also be used to study the emergence of behavior---such as associative learning---and to perform controlled experiments to test various evolutionary theories~\citep{pontes2020Evolutionary,moreno2022Exploring,ferguson2023Potentiating}.

Recent work, often stemming from the reinforcement learning (RL) community~\citep{johnson2016malmo,charity2023amorphous,rutherford2023jaxmarl,matthews2024craftax}, contains some elements of ALife simulations, and open-endedness~\citep{stanley2017open, leibo2019autocurricula, clune2019ai} in general.
One example is Neural MMO~\citep{suarez2019neural}, which defines a world of agents competing for a finite amount of resources. These agents are generally trained with RL, instead of competing via natural selection. XLand~\citep{team2021open} similarly defines an expressive collection of multi-agent environments, although much of the interesting variation comes in the form of tasks and worlds rather than agents. 
However, these environments are generally posed as relatively static learning environments for RL agents instead of environments to study evolutionary and open-ended behavior.

Other works that are more in the spirit of traditional ALife simulations include MineLand~\citep{yu2024mineland}, which defines a cooperative multi-agent world in Minecraft and the work by \citet{park2023generative}, which simulates a human-like community using pre-trained language models to interact between agents. Lenia \citep{chan2018lenia} is a cellular automaton that can result in the evolution of a diverse range of creatures with lifelike properties from low-level building blocks.

Recent hardware-accelerated ALife works include Alien \citep{alien}, which simulates agents that can evolve different morphologies in a particle-based system, and Biomaker CA \citep{randazzo2023biomaker} which evolves artificial plants in a grid-based world. While both are powerful simulations, their focus is more on the evolution of lower-level dynamics as opposed to higher-level agentic \textit{behaviors}.

\section{Conclusion and Future Work}
\looseness=-1 We introduce \name, where agents must survive and evolve within a changing world, with 
programmable robots affording unbounded complexity.
We believe \name could be used to explore alternative routes to technological and cultural advancements, such as the evolution of mathematics. Using \name to perform hypothesis-driven study of important questions in evolution would be a fruitful avenue for future work. For instance, we could change various hyperparameters of the simulation, or disable components such as communication, to see what effect these changes have on the final outcomes.
We also note that the simulation results can vary wildly when changing the initial conditions, so investigating the effects of different initial settings would also be valuable.

\name was designed at a higher level of abstraction than many comparable ALife simulations, completely ignoring the low-level aspects of physics, chemistry and control. This comes from our basic assumption that agents do not need to first evolve basic control and perception to be able to develop higher-level reasoning.
In particular, we take an \textit{anthropocentric} view of evolution, completely overlooking early evolutionary advancements in control, perception, and morphology and instead focusing entirely on human evolution.
Furthermore, we are potentially limited by having each agent be physically identical, where the only difference between agents is their behavior. Therefore, \name will not be able to examine physical speciation and different morphology-based niches; however, this is not our goal, and there are several prior simulations that are more well-suited to this~\citep{sims1994Evolving,silveira1998modeling,spector2007division,bessonov2015morphology,alien}.

There are several promising avenues for future work, such as using other parameterizations to control the agents' behavior, e.g., VSML~\citep{kirsch2021Meta}, where a small number of parameters can define an entire \textit{learning algorithm}, potentially being better suited to evolutionary methods. Another alternative would be to evolve a reinforcement learning objective~\citep{lu2022discovered, jackson2024discovering} or reward~\citep{sapora2023evil} function.
Another direction would be to investigate which features of the environment could encourage agents to interact more with the robots; for instance, adding near-extinction events that could encourage agents to store information to be used by future generations. Developing improved quantitative metrics to measure the interaction with the robots would also be promising; for instance, computing the Kolmogorov complexity of the system as a whole~\citep{kolmogorov1965three}. Moreover, investigating the altruistic or selfish nature of agents' behavior within a resource-constrained world would also be interesting~\citep{perolat2017multi,lupu2020gifting}.

Ultimately, we believe \name's large amount of potential complexity provides a good testbed for investigating the emergence of higher-level features, such as culture, language, and mathematics. Being able to interact with complex computational tools could lead to particularly interesting results when the simulation is run for long enough.

\footnotesize
\bibliographystyle{apalike}
\bibliography{example} 
\end{document}